\documentclass[final]{cvpr}

\usepackage{times}
\usepackage{amsmath,graphicx}
\usepackage{colortbl}
\definecolor{shadow}{cmyk}{.15,0,0,0}
\usepackage{amssymb}
\usepackage{array}
\usepackage{booktabs}
\usepackage{multirow}
\usepackage{graphicx}
\usepackage{color}
\usepackage{float}
\usepackage{stfloats}
\usepackage{stmaryrd}
\usepackage{subfig}
\usepackage{makecell}
\usepackage{threeparttable}
\usepackage{enumitem}
\usepackage{enumerate}
\usepackage[pagebackref=true,breaklinks=true,letterpaper=true,colorlinks,bookmarks=false]{hyperref}

\begin{document}

\title{Symmetric Parallax Attention for Stereo Image Super-Resolution}
\author{Yingqian Wang\thanks{Yingqian Wang and Xinyi Ying contribute equally to this work and are co-first authors. $^{\dagger}$Corresponding author: Jungang Yang.} ,~~Xinyi Ying$^{*}$,~~Longguang Wang,~~Jungang Yang$^{\dagger}$,~~Wei An,~~Yulan Guo\\
National University of Defense Technology\\
\tt\small \{wangyingqian16, yingxinyi18, yangjungang\}@nudt.edu.cn}

\maketitle

\begin{abstract}
   Although recent years have witnessed the great advances in stereo image super-resolution (SR), the beneficial information provided by binocular systems has not been fully used. Since stereo images are highly symmetric under epipolar constraint, in this paper, we improve the performance of stereo image SR by exploiting symmetry cues in stereo image pairs. Specifically, we propose a symmetric bi-directional parallax attention module (biPAM) and an inline occlusion handling scheme to effectively interact cross-view information. Then, we design a Siamese network equipped with a biPAM to super-resolve both sides of views in a highly symmetric manner. Finally, we design several illuminance-robust losses to enhance stereo consistency. Experiments on four public datasets demonstrate the superior performance of our method. Source code is available at \url{https://github.com/YingqianWang/iPASSR}.
\end{abstract}

\section{Introduction}\label{introduction}
   With recent advances in stereo vision, dual cameras are commonly adopted in mobile phones and autonomous vehicles. Using the complementary information (i.e., cross-view information) provided by binocular systems, the resolution of image pairs can be enhanced. However, it is challenging to achieve good performance in stereo image super-resolution (SR) due to the following issues: \textit{\textbf{1) Varying parallax.}} Objects at different depths have different disparity values and thus locate at different positions along the horizontal epipolar line. It is challenging to capture reliable stereo correspondence and effectively integrate cross-view information for stereo image SR.  \textit{\textbf{2) Information incorporation.}} Since context information within a single view (i.e., intra-view information) is crucial and contributes to stereo image SR in a different manner, it is important but challenging to fully incorporate both intra-view and cross-view information. \textit{\textbf{3) Occlusions \& boundaries.}} In occlusion and boundary areas, pixels in one view cannot find their correspondence in the other view. In this case, only intra-view information is available for stereo image SR. It is challenging to fully use cross-view information in non-occluded regions while maintaining promising performance in occluded regions.

  Recently, several methods have been proposed to address the above issues. Wang \etal \cite{PAM,PASSRnet} addressed the varying parallax issue by proposing a parallax attention module (PAM), and developed a \textit{PASSRnet} for stereo image SR. Ying \etal \cite{SAM} addressed the information incorporation issue by equipping several stereo attention modules (SAMs) to the pre-trained single image SR (SISR) networks. Song \etal \cite{SPAM} addressed the occlusion issue by checking stereo consistency using disparity maps regressed by parallax attention maps. Although continuous improvements have been achieved, the inherent correlation within stereo image pairs are still under exploited, which hinders the performance of stereo image SR.
%
% \begin{figure}
% \centering
% \includegraphics[width=8.2cm]{Figs/Thumbnail.pdf}
%\vspace{-0.1cm}
% \caption{Results achieved by different methods on scene \textit{Test\_0002} \cite{Flickr} for 4$\times$SR. Our iPASSR recovers more faithful and stereo-consistent details than other methods.}\label{FigThumbnail}
% \vspace{-0.2cm}
% \end{figure}

  Since super-resolving left and right images are highly symmetric, the inherent correlation within an image pair can be fully used by exploiting its symmetry cues. In this paper, we improve the performance of stereo image SR by exploiting symmetries on three levels. \textbf{\textit{1) On the module level}}, we design a symmetric bi-directional parallax attention module (biPAM) to interact cross-view information. With our biPAM, occlusion maps can be generated and used as a guidance for cross-view feature fusion. \textbf{\textit{2) On the network level}}, we propose a Siamese network equipped with our biPAM to super-resolve both left and right images. Experimental results demonstrate that jointly super-resolving both sides of views can better exploit the correlation between stereo images and is contributive to SR performance. \textbf{\textit{3) On the optimization level}}, we exploit symmetry cues by designing several bilateral losses. Our proposed losses can enforce stereo consistency and is robust to illuminance changes between stereo images. We perform extensive ablation studies to validate the effectiveness of our method. Comparative results on the \textit{KITTI 2012} \cite{K12}, \textit{KITTI 2015} \cite{K15}, \textit{Middlebury} \cite{Mid} and \textit{Flickr1024} \cite{Flickr} datasets have demonstrated the competitive performance of our method as compared to many state-of-the-art SR methods.

 \begin{figure*}
 \centering
 \includegraphics[width=17cm]{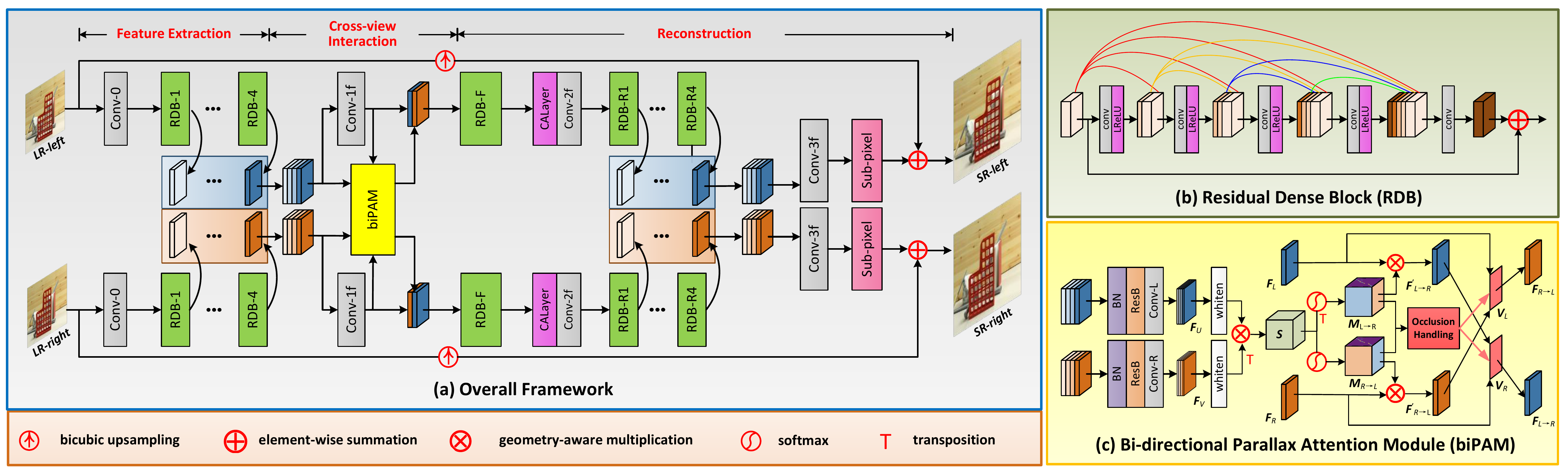}
  \vspace{-0.2cm}
 \caption{An overview of our \textit{iPASSR} network.}\label{FigNetwork}
 \vspace{-0.05cm}
 \end{figure*}

  Our proposed method is named \textbf{\textit{iPASSR}} since it is an improved version of our previous \textit{PASSRnet} \cite{PAM,PASSRnet}. The contributions of this paper are as follows: \textbf{\textit{1)}} We propose to exploit symmetry cues for stereo image SR. Different from \textit{PASSRnet}, our \textit{iPASSR} can super-resolve both sides of views within a single inference. \textbf{\textit{2)}} We develop a symmetric and bi-directional parallax attention module. Compared to PAMs in \cite{PAM,PASSRnet}, our biPAM is more compact and can effectively handle occlusions. \textbf{\textit{3)}}  As demonstrated in the experiments, our \textit{iPASSR} can achieve significant performance improvements over \textit{PASSRnet} with a comparable model size.

  The rest of this paper is organized as follows. In Section~\ref{SecRelatedWork}, we briefly review the related work. In Section~\ref{SecMethod}, we introduce our proposed method including network architecture, occlusion handling scheme, and loss functions. Experimental results are presented in Section~\ref{SecExperiments}. Finally, we conclude this paper in Section~\ref{SecConclusion}.

  \section{Related Work}\label{SecRelatedWork}
  \textbf{Single Image SR.} SISR is a long-standing problem and has been investigated for decades. Recently, deep learning-based SISR methods have achieved promising performance in terms of both reconstruction accuracy \cite{USRnet,DPSR,SMSR,DASR} and visual quality \cite{SRGAN,ESRGAN,GLEAN,BSRGAN}. Dong \etal \cite{SRCNN2014} proposed the first CNN-based SR network named \textit{SRCNN} to reconstruct high-resolution (HR) images from low-resolution (LR) inputs. Kim \etal \cite{VDSR} proposed a very deep network (\textit{VDSR}) with 20 layers to improve SR performance. Afterwards, SR networks became increasingly deep and complex, and thus more powerful in intra-view information exploitation. Lim \etal \cite{EDSR} proposed an enhanced deep SR network (\textit{EDSR}) using both local and global residual connections. Zhang \etal \cite{RDN,RDN+} combined residual connection with dense connection, and proposed residual dense network (\textit{RDN}) to fully exploit hierarchical feature representations. More recently, the performance of SISR has been further improved by \textit{RCAN} \cite{RCAN}, \textit{RNAN} \cite{RNAN} and \textit{SAN} \cite{SAN}.

  \textbf{Stereo Image SR.} Compared to SISR which exploits context information within only one view, stereo image SR aims at using the cross-view information provided by stereo images. Jeon \etal \cite{StereoSR} proposed a network named \textit{StereoSR} to learn a parallax prior by jointly training two cascaded sub-networks. The cross-view information is integrated by concatenating the left image and a stack of right images with different pre-defined shifts. Wang \etal \cite{PAM,PASSRnet} proposed a parallax attention module to learn stereo correspondence with a global receptive field along the epipolar line. Ying \etal \cite{SAM} proposed a stereo attention module and embedded it into pre-trained SISR networks for stereo image SR. Song \etal \cite{SPAM} combined self-attention with parallax attention for stereo image SR. Furthermore, stereo consistency was addressed by using disparity maps regressed from parallax attention maps. Yan \etal \cite{DASSR} proposed a domain adaptive stereo SR network (\ie, \textit{DASSR}). Specifically, they first explicitly estimated disparities using a pretrained stereo matching network \cite{StereoNet} and then warped views to the other side to incorporate cross-view information. More recently, Xu \etal \cite{BSSRnet} incorporated bilateral grid processing into CNNs and proposed a \textit{BSSRnet} for stereo image SR.

 \section{Method}\label{SecMethod}

 In this section, we introduce our method in details. We first introduce  the architecture of our network in Section \ref{SecNetwork}, then describe the inline occlusion handling scheme in Section \ref{SecOcclusion}. Finally, we present the losses in Section \ref{SecLosses}.

 \subsection{Network Architecture}\label{SecNetwork}
 Our network takes a pair of LR RGB stereo images $\mathbf{I}_\text{L}^\text{input}$ and $ \mathbf{I}_\text{R}^\text{input}$ as its inputs to generate HR RGB stereo images $\mathbf{I}_\text{L}^\text{SR}$ and $\mathbf{I}_\text{R}^\text{SR}$. As shown in Fig.~\ref{FigNetwork}(a), our network is highly symmetric and the weights of its left and right branches are shared. Given LR input stereo images, our network sequentially performs \textbf{\textit{feature extraction}}, \textbf{\textit{cross-view interaction}}, and \textbf{\textit{reconstruction}}.

 \textbf{Feature Extraction.}
 In our feature extraction module, input stereo images $\mathbf{I}_\text{L}^\text{input}, \mathbf{I}_\text{R}^\text{input} \in\mathbb{R}^{H \times W \times 3}$ are first fed to a convolution layer (\ie, \textit{Conv-0}) to generate initial features $\mathbf{F}_\text{L}^{0}, \mathbf{F}_\text{R}^{0} \in\mathbb{R}^{H \times W \times 64}$, which are then fed to 4 cascaded residual dense blocks (RDBs)\footnote{The insights of using RDBs for feature extraction are two-folds: \textbf{First}, RDB can generate features with large receptive fields and dense sampling rates, which are beneficial to stereo correspondence estimation.   \textbf{Second}, RDB can fully use features from all the layers via local dense connection. The generated hierarchical features are beneficial to SR performance.} for deep feature extraction. As shown in Fig.~\ref{FigNetwork}(b),  4 convolutions with a growth rate of 24 are used within each RDB to achieve dense feature representation. Note that, features from all the layers in an RDB are concatenated and fed to a 1$\times$1 convolution to generate fused features for local residual connection.

 \textbf{Cross-view Interaction.}
 We propose a bi-directional parallax attention module (biPAM) to interact cross-view information of stereo features. Since hierarchical feature representation is beneficial to stereo correspondence learning \cite{PAM}, we form the inputs of our biPAM by concatenating the output features of each RDB in our feature extraction module. As shown in Fig.~\ref{FigNetwork}(c), the input stereo features are first fed to a batch-normalization (BN) layer and a transition residual block (\ie, \textit{ResB}), and then separately fed to 1$\times$1 convolutions to generate $\mathbf{F}_\text{U}$, $\mathbf{F}_\text{V} \in \mathbb{R}^{H \times W \times 64}$. To achieve disentangled pairwise parallax attention, we follow \cite{yin2020disentangled} and feed $\mathbf{F}_\text{U}$ and $\mathbf{F}_\text{V}$ to a whiten layer to obtain normalized features $\mathbf{F}'_\text{U}$ and $\mathbf{F}'_\text{V}$ according to
 \begin{equation}\label{EqWhiten1}
\mathbf{F}'_\text{U}(h,w,c) = \mathbf{F}_\text{U}(h,w,c) - \frac{1}{W} \sum\nolimits_{i=1}^{W}\mathbf{F}_\text{U}(h,i,c),
 \end{equation}
 \begin{equation}\label{EqWhiten2}
\mathbf{F}'_\text{V}(h,w,c) = \mathbf{F}_\text{V}(h,w,c) - \frac{1}{W} \sum\nolimits_{i=1}^{W}\mathbf{F}_\text{V}(h,i,c).
 \end{equation}

 To generate left and right attention maps, $\mathbf{F}'_\text{V}$ is first transposed to ${\mathbf{F}'_\text{V}}^\text{T}\in\mathbb{R}^{H \times 64 \times W}$, and then batch-wisely multiplied (see Section~\ref{SecOcclusion}) with $\mathbf{F}'_\text{U}$ to produce an initial score map $\mathbf{S}\in\mathbb{R}^{H \times W \times W}$. Then, softmax normalization is applied to $\mathbf{S}$ and $\mathbf{S}^\text{T}$ along their last dimension to generate attention maps $\mathbf{M}_\text{R$\rightarrow$L}$ and $\mathbf{M}_\text{L$\rightarrow$R}$, respectively. To achieve cross-view interaction, both left and right features (generated by \textit{Conv-1f} in Fig.~\ref{FigNetwork}(a)) need to be converted to the other side by taking a batch-wise matrix multiplication with the corresponding attention maps, \ie,
  \begin{equation}
 \mathbf{F'}_\text{R$\rightarrow$L} =  \mathbf{M}_\text{R$\rightarrow$L} \otimes  \mathbf{F}_\text{R},
 \end{equation}
 \begin{equation}
 \mathbf{F'}_\text{L$\rightarrow$R} =  \mathbf{M}_\text{L$\rightarrow$R} \otimes  \mathbf{F}_\text{L},
 \end{equation}
 where $\otimes$ denotes the batch-wise matrix multiplication.

 To avoid unreliable correspondence in occlusion and boundary regions, we propose an inline occlusion handling scheme to calculate valid masks $\mathbf{V}_\text{L}$ and $\mathbf{V}_\text{R}$. The final converted features $\mathbf{F}_\text{R$\rightarrow$L}$ and $\mathbf{F}_\text{L$\rightarrow$R}$ can be obtained by
 \begin{equation}\label{EqFuse1}
 \mathbf{F}_\text{R$\rightarrow$L} =  \mathbf{V}_\text{L} \odot  \mathbf{F'}_\text{R$\rightarrow$L} + (\mathbf{1} - \mathbf{V}_\text{L}) \odot  \mathbf{F}_\text{L},
 \end{equation}
  \begin{equation}\label{EqFuse2}
 \mathbf{F}_\text{L$\rightarrow$R} =  \mathbf{V}_\text{R} \odot  \mathbf{F'}_\text{L$\rightarrow$R} + (\mathbf{1} - \mathbf{V}_\text{R}) \odot  \mathbf{F}_\text{R},
 \end{equation}
 where $\odot$ represents element-wise multiplication. Note that, values in $\mathbf{V}_\text{L}$ and $\mathbf{V}_\text{R}$ range from 0 (occluded) to 1 (non-occluded). According to Eqs.~(\ref{EqFuse1}) and (\ref{EqFuse2}), occluded regions of converted features (\ie, $\mathbf{F}_\text{R$\rightarrow$L}$,  $\mathbf{F}_\text{L$\rightarrow$R}$) can be filled with the corresponding features from the target view (\ie, $\mathbf{F}_\text{L}$, $\mathbf{F}_\text{R}$), resulting in continuous spatial distributions.

  \textbf{Reconstruction.}
 Similar to the feature extraction module, we use RDB as the basic block in our reconstruction module. Taking the left branch as an example, $\mathbf{F}_{\text{R}\rightarrow \text{L}}$ is first concatenated with $\mathbf{F}_\text{L}$ and then fed to an RDB (\ie, \textit{RDB-F}) for initial feature fusion. The output feature $\mathbf{F}_\text{L}^\text{init\_f}\in\mathbb{R}^{H \times W \times 128}$ is then fed to a channel attention layer (\ie, \textit{CALayer} \cite{RCAN}) and a convolution layer (\ie, \textit{Conv-2f}) to produce the final fused feature $\mathbf{F}_\text{L}^\text{f}\in\mathbb{R}^{H \times W \times 64}$. Afterwards, $\mathbf{F}_\text{L}^\text{f}$ is fed to 4 cascaded RDBs, a convolution layer (\ie, \textit{Conv-3f}), and a sub-pixel layer \cite{PixelShuffle} to generate the super-resolved left image $\mathbf{I}_\text{L}^\text{SR}$.

 \begin{figure}
 \centering
 \includegraphics[width=7.5cm]{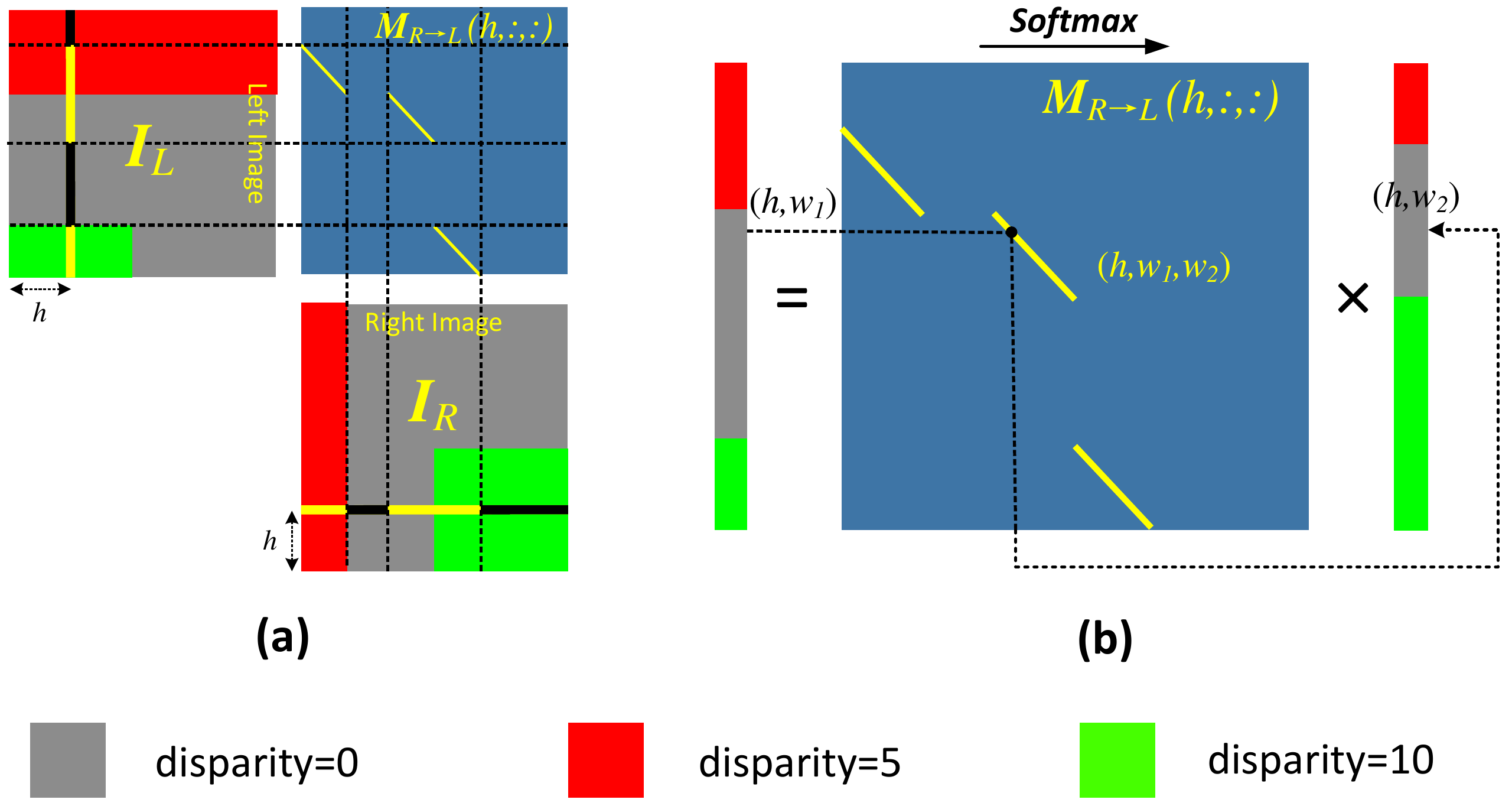}
  \vspace{-0.1cm}
 \caption{A toy example to depict the stereo correspondence. The gray, red, and green regions in $\mathbf{I}_\text{L}$ and $\mathbf{I}_\text{R}$ denote objects with a disparity of $0$, $5$, and $10$ pixels, respectively. For simplicity, only a profile of $\mathbf{M}_\text{R$\rightarrow$L}$ at height $h$ is visualized, which corresponds to the regions marked by yellow strokes in (a). Occlusions (colored in black on the strokes) are implicitly encoded in the attention maps as empty intervals. (b) The right stroke can be converted into the left side by multiplying it with $\mathbf{M}_\text{R$\rightarrow$L}$.}\label{FigToy}
 \vspace{-0.1cm}
 \end{figure}
 \subsection{Inline Occlusion Handling Scheme}\label{SecOcclusion}
 By using biPAM, the stereo correspondence can be generated in a symmetric manner. More importantly, the occlusions can be derived by checking the stereo consistency using the attention maps $\mathbf{M}_\text{R$\rightarrow$L}$ and $\mathbf{M}_\text{L$\rightarrow$R}$.

 Here, we use a toy example in Fig.~\ref{FigToy} to illustrate how occlusions are implicitly encoded in the parallax attention maps. Given a pair of stereo images $\mathbf{I}_\text{L}$ and $\mathbf{I}_\text{R}\in \mathbb{R}^{H\times W}$, parallax attention maps $\mathbf{M}_\text{R$\rightarrow$L},\mathbf{M}_\text{L$\rightarrow$R} \in \mathbb{R}^{H\times W\times W}$ can be generated. As illustrated in Fig.~\ref{FigToy}(a),  we visualize a profile of $\mathbf{M}_\text{R$\rightarrow$L}$ at height $h$ (\ie, $\mathbf{M}_{\text{R$\rightarrow$L}}(h,:,:)$), which corresponds to the yellow strokes in the left and right images. Note that, black strokes represent occluded regions. It can be observed from Fig.~\ref{FigToy}(a) that: \textbf{\textit{1)}} Occlusions occur near object edges where the depth values change suddenly, or occur near image boundaries (more specifically, left boundary of the left view and right boundary of the right view).  \textbf{\textit{2)}} The occluded regions correspond to the empty intervals in the attention maps since their counterparts in the other view are unavailable. These two observations demonstrate that occlusions are implicitly encoded in the parallax attention maps and can be calculated by checking the cycle consistency using $\mathbf{M}_\text{R$\rightarrow$L}$ and $\mathbf{M}_\text{L$\rightarrow$R}$. Specifically, the right image can be converted into the left side according to $\mathbf{I}_\text{R$\rightarrow$L} = \mathbf{M}_\text{R$\rightarrow$L} \otimes \mathbf{I}_\text{R}$, where $\otimes$ represents the batch-wise matrix multiplication. As shown in Fig.~\ref{FigToy}(b), the product of a slice of the right image (\ie, $\mathbf{I}_\text{R}(h,:)$) and the corresponding profile of the attention map (\ie, $\mathbf{M}_\text{R$\rightarrow$L}(h,:,:)$) determines the slice of the converted left image at the same height (\ie, $\mathbf{I}_\text{R$\rightarrow$L}(h,:)$). All these resulting slices are concatenated to produce $\mathbf{I}_\text{R$\rightarrow$L}$.

Note that, softmax normalization has been performed along the third dimension of $\mathbf{M}_\text{R$\rightarrow$L}$ and $\mathbf{M}_\text{L$\rightarrow$R}$. Therefore, $\mathbf{M}_\text{R$\rightarrow$L}(h,w_1,w_2)$ can be considered as the matching possibility between $\mathbf{I}_\text{R}(h,w_2)$ and $\mathbf{I}_\text{L}(h,w_1)$. Furthermore, the possibility that $\mathbf{I}_\text{L}(h,w_1)$ is first converted to $\mathbf{I}_\text{R}$ and then re-converted to $\mathbf{I}_\text{L}(h,w_1)$ can be calculated according to
\begin{equation}\label{EqPl}
\mathbf{P}_\text{L}(h, w_1) = \sum_{w_{2}=1}^{W} \mathbf{M}_\text{R$\rightarrow$L}(h,w_1,w_2) \cdot \mathbf{M}_\text{L$\rightarrow$R}(h,w_2,w_1).
\end{equation}

Note that,  $\mathbf{P}_\text{L}(h, w_1)$ is close to 0 if point $(h, w_1)$ is occluded in the right view. Consequently, $\mathbf{P}_\text{L}$ can be used to represent occlusions in the left image. Due to noise and rectification errors in stereo images, we relax the constraint in Eq.~\ref{EqPl} by $\pm2$ pixels in this work:
\begin{equation}
\begin{split}
\mathbf{P}'_\text{L}(h, w_1) = \sum_{\delta=-2}^{2} \sum_{w_{2}=1}^{W} &\mathbf{M}_\text{R$\rightarrow$L}(h,w_1 + \delta, w_2) \cdot \\
& \mathbf{M}_\text{L$\rightarrow$R}(h,w_2,w_1).
\end{split}
\end{equation}

\begin{figure}
 \centering
 \includegraphics[width=8.2cm]{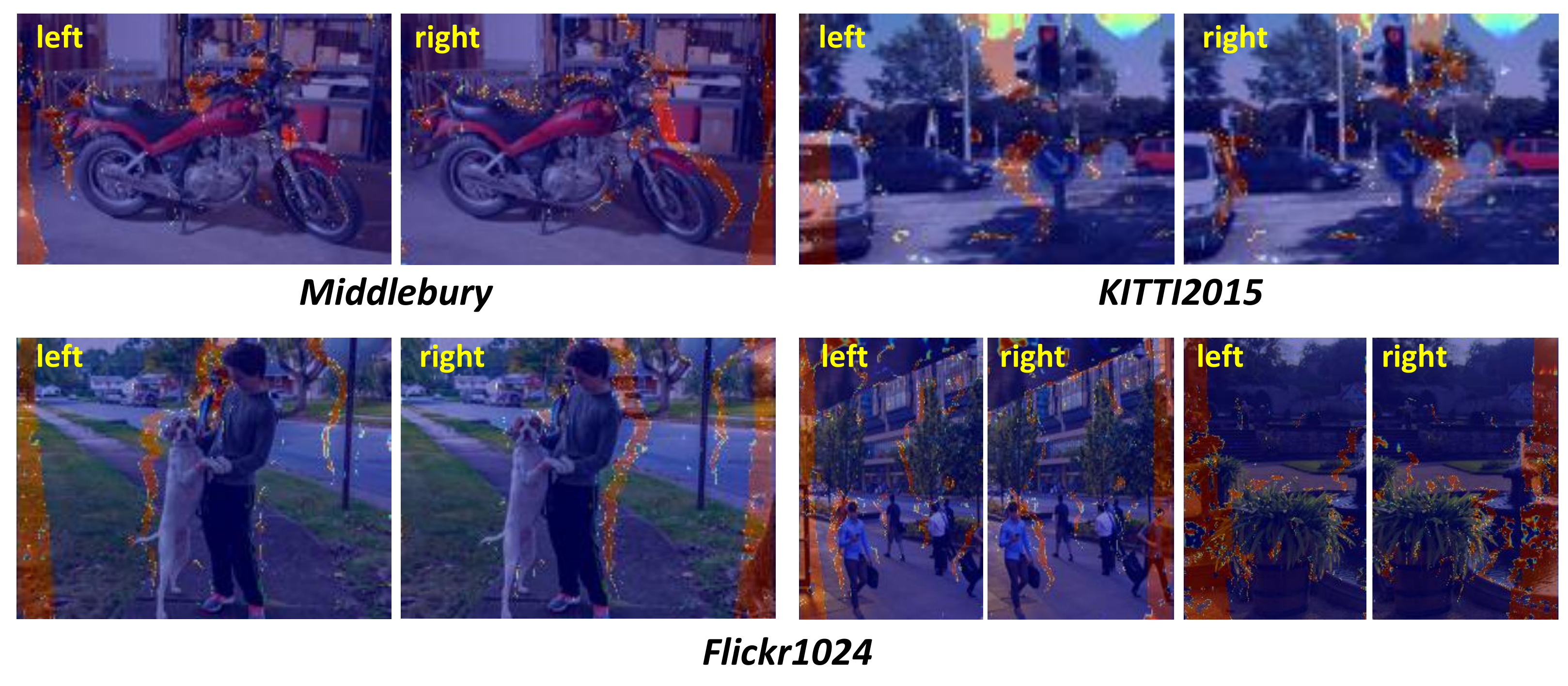}
  \vspace{-0.2cm}
 \caption{An illustration of valid masks generated by our occlusion handling scheme. Red regions have small values and represent heavy occlusions.}\label{FigMask}
 \vspace{-0.2cm}
 \end{figure}

To maintain training stability, the left valid mask $\mathbf{V}_\text{L}$ is calculated according to $\mathbf{V}_\text{L} = \text{tanh}(\tau\mathbf{P}'_\text{L})$, where $\tau$ was empirically set to 5 in our implementation. The right valid mask $\mathbf{V}_\text{R}$ can be generated following a similar way. Figure~\ref{FigMask} shows some examples of the generated valid masks.

 \subsection{Losses}\label{SecLosses}
The overall loss function of our network is defined as:
\begin{equation}
\mathcal{L} = \mathcal{L}_\text{SR} + \lambda (\mathcal{L}_\text{photo}^\text{res} + \mathcal{L}_\text{cycle}^\text{res} + \mathcal{L}_\text{smooth} + \mathcal{L}_\text{cons}^\text{res}),
\end{equation}
where $\mathcal{L}_\text{SR}$, $\mathcal{L}_\text{photo}^\text{res}$, $\mathcal{L}_\text{cycle}^\text{res}$, $\mathcal{L}_\text{smooth}$, and $\mathcal{L}_\text{cons}^\text{res}$ represent SR loss, residual photometric loss, residual cycle loss, smoothness loss, and residual stereo consistency loss, respectively. $\lambda$ represents the weight of the regularization term and was empirically set to 0.1 in this work. The \textit{SR loss} is defined as the $L_{1}$ distance between the super-resolved and groundtruth stereo images:
\begin{equation}
\mathcal{L}_\text{SR} = \parallel \mathbf{I}_\text{L}^\text{SR} - \mathbf{I}_\text{L}^\text{HR} \parallel_1 +  \parallel \mathbf{I}_\text{R}^\text{SR} - \mathbf{I}_\text{R}^\text{HR} \parallel_{1},
\end{equation}
where $\mathbf{I}_\text{L}^\text{SR}$ and $\mathbf{I}_\text{R}^\text{SR}$ represent the super-resolved left and right images, $\mathbf{I}_\text{L}^\text{HR}$ and $\mathbf{I}_\text{R}^\text{HR}$ represent their groundtruth HR images.

\begin{figure}
\centering
\includegraphics[width=8.3cm]{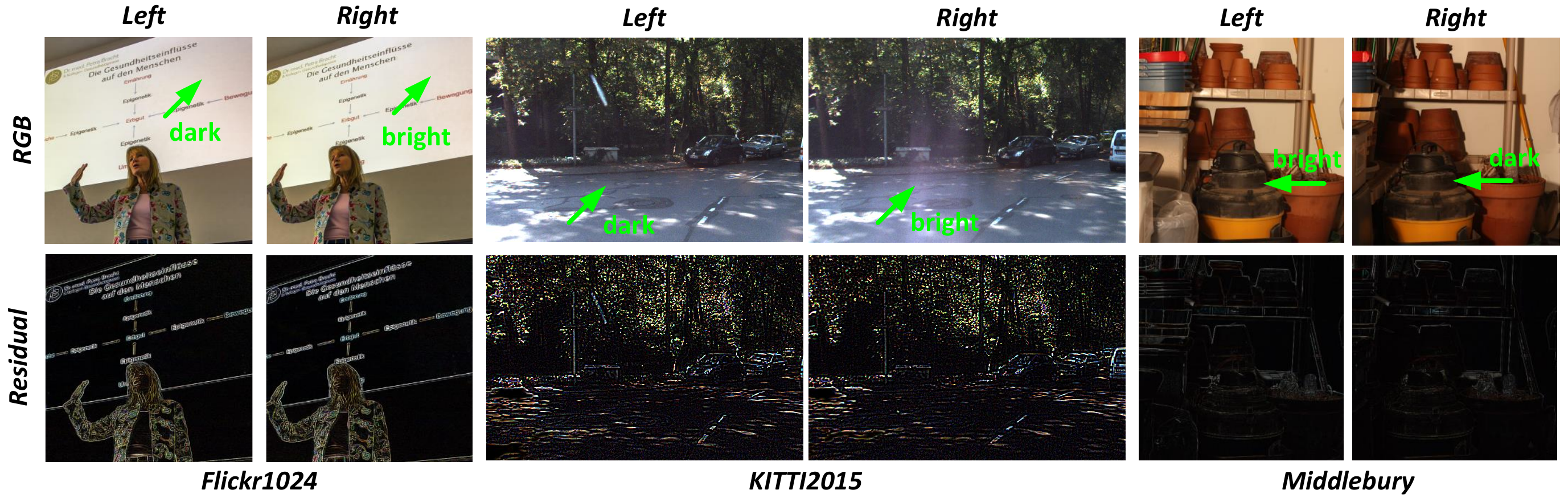}
\vspace{-0.4cm}
\caption{An illustration of illuminance changes in stereo image pairs. View our \href{https://wyqdatabase.s3-us-west-1.amazonaws.com/iPASSR_illuminance_change.mp4}{demo video} for better visualization.}\label{FigIllum}
\vspace{-0.1cm}
\end{figure}

Due to exposure difference and non-Lambertain surfaces, the illuminance intensity between stereo images can vary significantly (see Fig.~\ref{FigIllum}). In these cases, the photometric loss and cycle loss used in \cite{PASSRnet,PAM,SAM,SPAM} can lead to a mismatch problem. To handle this problem, we calculate these losses using residual images to improve their robustness to illuminance changes. Specifically, we introduce $\mathbf{X}_\text{L} =\left| \mathbf{I}_\text{L}^\text{HR} - \mathbf{I}_\text{L}^\text{LR}\uparrow \right| \downarrow$ and $\mathbf{X}_\text{R} = \left| \mathbf{I}_\text{R}^\text{HR} - \mathbf{I}_\text{R}^\text{LR}\uparrow \right| \downarrow$, where $\uparrow$ and $\downarrow$ represent bicubic upsampling and downsampling, and $\mathbf{X}_\text{L} $ and $\mathbf{X}_\text{R} $ represent the absolute values of the left and right residual images, respectively. Consequently, the residual photometric loss and residual cycle consistency loss can be formulated as:
\begin{equation}
\begin{split}
\mathcal{L}_\text{photo}^\text{res}& = \parallel V_\text{L} \odot (\mathbf{X}_\text{L} - \mathbf{M}_\text{R$\rightarrow$ L}\otimes \mathbf{X}_\text{R}) \parallel_1 \\
 & + \parallel V_\text{R} \odot (\mathbf{X}_\text{R} - \mathbf{M}_\text{L$\rightarrow$R} \otimes \mathbf{X}_\text{L}) \parallel_1,
\end{split}
\end{equation}
\begin{equation}
\begin{split}
\mathcal{L}_\text{cycle}^\text{res} & =\parallel V_\text{L} \odot (\mathbf{X}_\text{L} - \mathbf{M}_\text{R$\rightarrow$L}\otimes \mathbf{M}_\text{L$\rightarrow$R}\otimes \mathbf{X}_\text{L}) \parallel_1 \\
 & + \parallel V_\text{R} \odot (\mathbf{X}_\text{R} - \mathbf{M}_\text{L$\rightarrow$R}\otimes \mathbf{M}_\text{R$\rightarrow$L}\otimes \mathbf{X}_\text{R}) \parallel_1.
\end{split}
\end{equation}

Residual photometric and cycle losses introduce two benefits. \textbf{First}, since illuminance components can be eliminated, more consistent and illuminance-robust stereo correspondence can be learned by our biPAM. \textbf{Second}, since residual images mainly contain high-frequency components, our biPAM can pay more attention to texture-rich regions, which is contributive to SR performance.

Apart from the aforementioned losses, we also employ smoothness loss to encourage smoothness in correspondence space. That is,
\begin{equation}
\begin{split}
\mathcal{L}_\text{smooth}& = \sum_{\mathbf{M}}\sum_{i,j,k} ( \parallel \mathbf{M}(i,j,k) - \mathbf{M}(i+1,j,k) \parallel_1 \\
 & + \parallel \mathbf{M}(i,j,k) - \mathbf{M}(i,j+1,k+1) \parallel_1),
\end{split}
\end{equation}
where $\mathbf{M}\in\{\mathbf{M}_\text{R$\rightarrow$L},\mathbf{M}_\text{L$\rightarrow$R}\}$. Here, $\parallel\mathbf{M}_\text{R$\rightarrow$L}(i,j,k) - \mathbf{M}_\text{R$\rightarrow$L}(i+1,j,k) \parallel_1$ enforces the correspondence between $\mathbf{I}_\text{R}(i+1,k)$ and $\mathbf{I}_\text{L}(i+1,j)$ to be close to the correspondence between $\mathbf{I}_\text{R}(i,k)$ and $\mathbf{I}_\text{L}(i,j)$.

Finally, we introduce residual stereo consistency loss to achieve stereo consistency between super-resolved left and right images. Specifically, the LR residuals between super-resolved images and groundtruth images are calculated according to
$\mathbf{Y}_\text{L} = \left|\mathbf{I}_\text{L}^\text{HR} - \mathbf{I}_\text{L}^\text{SR} \right|\downarrow$ and
$\mathbf{Y}_\text{R} = \left| \mathbf{I}_\text{R}^\text{HR} - \mathbf{I}_\text{R}^\text{SR} \right| \downarrow$, respectively, and the residual stereo consistency loss is defined as
\begin{equation}
\begin{split}
\mathcal{L}_\text{cons}^\text{res}& = \parallel V_\text{L} \odot (\mathbf{Y}_\text{L} - \mathbf{M}_\text{R$\rightarrow$L}\otimes \mathbf{Y}_\text{R}) \parallel_1 \\
 & + \parallel V_\text{R} \odot (\mathbf{Y}_\text{R} - \mathbf{M}_\text{L$\rightarrow$R}\otimes \mathbf{Y}_\text{L}) \parallel_1.
\end{split}
\end{equation}

%Different from the photometric loss which enforces biPAM to learn accurate stereo correspondence, our residual stereo consistency loss enforces the super-resolved stereo images to be close to each other in non-occluded corresponding regions. To avoid training conflict, we canceled the gradient back propagation of $\mathbf{M}_{R\rightarrow L}$ and $\mathbf{M}_{L\rightarrow R}$ to ensure stable updating of super-resolved images.

\section{Experiments}\label{SecExperiments}
In this section, we first introduce the datasets and implementation details, then perform ablation studies to validate our design choices. Finally, we compare our \textit{iPASSR} to several state-of-the-art SISR and stereo image SR methods.

\subsection{Datasets and Implementation Details}
We used 800 images from the training set of \textit{Flickr1024} \cite{Flickr} and 60 images from \textit{Middlebury} \cite{Mid} as the training data. For images from the \textit{Middlebury} dataset, we followed \cite{StereoSR,PASSRnet,SAM} to perform bicubic downsampling with a factor of 2 to generate HR images. For test, we followed \cite{StereoSR,PASSRnet,SAM} to generate our test set by using 5 images from \textit{Middlebury} \cite{Mid}, 20 images from \textit{KITTI 2012} \cite{K12} and 20 images from \textit{KITTI 2015} \cite{K15}. Moreover, we used the test set of \textit{Flickr1024} \cite{Flickr} for additional evaluation. We used the bicubic downsampling approach to generate LR images. During the training phase, the generated LR images were cropped into patches of size $30\times90$ with a stride of 20, and their HR counterparts were cropped accordingly. These patches were randomly flipped horizontally and vertically for data augmentation.

Peak signal-to-noise ratio (PSNR) and structural similarity (SSIM) were used as quantitative metrics. To achieve fair comparison with \cite{StereoSR,PASSRnet,SAM}, we followed these methods to calculate PSNR and SSIM on the left views with their left boundaries (64 pixels) being cropped. Moreover, to comprehensively evaluate the performance of stereo image SR, we also report the average PSNR and SSIM scores on stereo image pairs (\ie, $\left(\textit{Left}+\textit{Right}\right)/2$) without any boundary cropping.

Our network was implemented in PyTorch on a PC with two Nvidia RTX 2080Ti GPUs. All models were optimized using the Adam method with $\beta_{1}=0.9$, $\beta_{2}=0.999$ and a batch size of 36. The initial learning rate was set to $2\times10^{-4}$ and reduced to half after every 30 epochs. The training was stopped after 80 epochs since more epochs do not provide further consistent improvement.

\subsection{Ablation Study}

\begin{table}
\centering
\footnotesize
\caption{Results achieved on the \textit{KITTI 2015} dataset by our method with different cross-view information incorporation schemes for $4\times$SR. Here, PSNR$/$SSIM of the cropped left views are reported.} \label{TabCross}
 \vspace{-0.2cm}
\begin{tabular}{|l|c|c|}
\hline
Models & Inputs & PSNR$/$SSIM \\
\hline
\textit{iPASSR with single input}           & Left           & 25.316$/$0.7753 \\
\textit{iPASSR with replicated inputs}   & Left-Left    & 25.400$/$0.7775 \\
\textit{Asymmetric iPASSR}                & Left-Right  & 25.548$/$0.7829 \\
\textit{iPASSR}                                   & Left-Right  & 25.615$/$0.7850 \\
\hline
\end{tabular}
 \vspace{-0.1cm}
\end{table}

\textbf{Cross-view information.}
We removed biPAM and retrained a single branch of our \textit{iPASSR} on the same training set as our original network. In addition, we also used pairs of replicated left images as inputs to directly perform inference using our original network. As shown in Table~\ref{TabCross}, the network trained with single images  (\ie, \textit{iPASSR with single input}) suffers a decrease of 0.299 dB in PSNR as compared to the original network. If replicated left images were used as inputs, the performance of the variant  (\ie, \textit{iPASSR with replicated inputs}) is also notably inferior to our original network. These results demonstrate the importance of cross-view information for stereo image SR.

\textbf{Siamese network architecture.}
We investigate the benefits introduced by our Siamese network architecture by retraining the network with stereo images as inputs but only super-resolving the left view (\ie, \textit{Asymmetric iPASSR}). It can be observed in Table~\ref{TabCross} that the PSNR score achieved by Asymmetric iPASSR is marginally lower than our \textit{iPASSR} (25.548 v.s. 25.615). That is because, the symmetric Siamese network structure can help to better exploit the cross-view information to improve the SR performance.

\begin{table}
 \caption{Results achieved on the \textit{KITTI 2015} dataset \cite{K15} by \textit{iPASSR} with different losses for $4\times$SR. ``\textit{Res}'' represents $\mathcal{L}_\text{photo}$, $\mathcal{L}_\text{cycle}$, and $\mathcal{L}_\text{cons}$ calculated on residual images. Here, PSNR$/$SSIM values of the cropped left views are reported.}\label{TabLoss}
  \vspace{-0.2cm}
 \centering
 \scriptsize
 \begin{tabular}{|cccccc|c|}
 \hline
 $\mathcal{L}_\text{SR}$ & $\mathcal{L}_\text{photo}$ & $\mathcal{L}_\text{smooth}$ & $\mathcal{L}_\text{cycle}$ & $\mathcal{L}_\text{cons}$ & \textit{Res} & PSNR$/$SSIM \\
 \hline
 $\checkmark$ &                       &                       &                       &                      &                       & 25.527$/$0.7827 \\
 \hline
 $\checkmark$ & $\checkmark$ &                       &                       &                      & $\checkmark$ & 25.535$/$0.7815 \\
 \hline
 $\checkmark$ & $\checkmark$ & $\checkmark$ &                       &                       & $\checkmark$ & 25.481$/$0.7795 \\
 \hline
  $\checkmark$ & $\checkmark$ &                      &  $\checkmark$  &                     & $\checkmark$ & 25.552$/$0.7839 \\
 \hline
 $\checkmark$ & $\checkmark$ & $\checkmark$ & $\checkmark$ &                       & $\checkmark$ & 25.570$/$0.7839 \\
 \hline
 $\checkmark$ & $\checkmark$ & $\checkmark$ & $\checkmark$ & $\checkmark$ &                       & 25.456$/$0.7775 \\
 \hline
 $\checkmark$ & $\checkmark$ & $\checkmark$ & $\checkmark$ & $\checkmark$ & $\checkmark$ & 25.615$/$0.7850 \\
\hline
 \end{tabular}
  \vspace{-0.0cm}
 \end{table}

\begin{figure}[t]
 \centering
 \includegraphics[width=8.2cm]{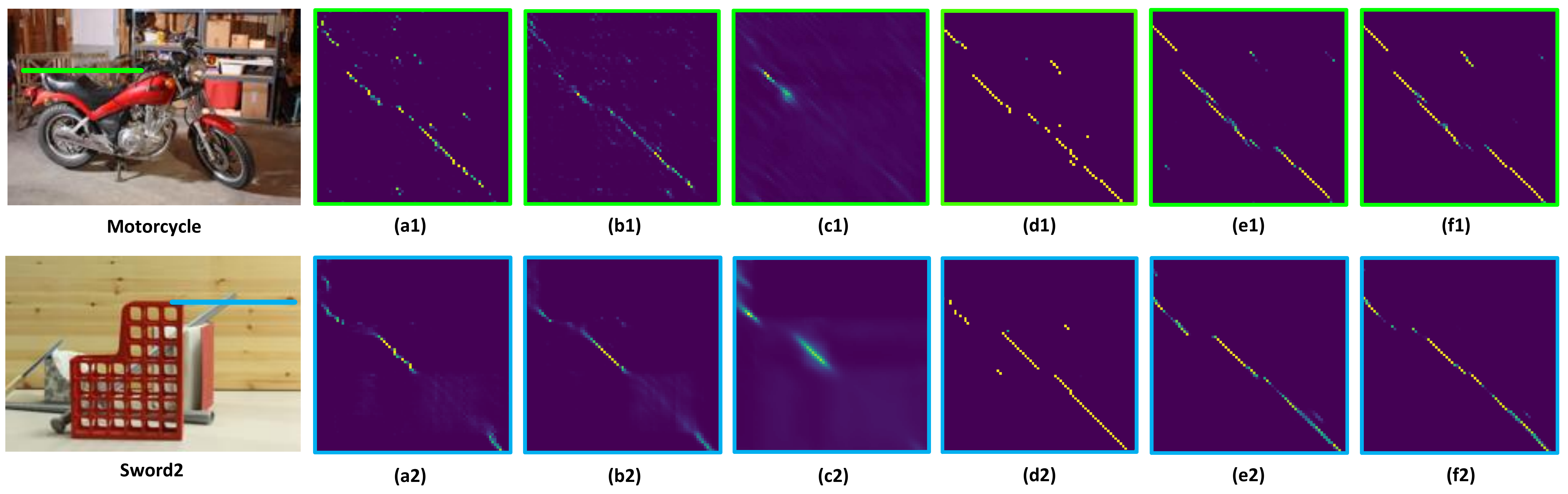}
  \vspace{-0.1cm}
 \caption{Visualization of attention maps generated by our \textit{iPASSR} trained with different losses. (a) $\mathcal{L}_\text{SR}$, (b) $\mathcal{L}_\text{SR}+ \lambda\mathcal{L}_\text{photo}^\text{res}$, (c) $\mathcal{L}_\text{SR} + \lambda(\mathcal{L}_\text{photo}^\text{res} + \mathcal{L}_\text{smooth})$, (d) $\mathcal{L}_\text{SR} + \lambda(\mathcal{L}_\text{photo}^\text{res} + \mathcal{L}_\text{cycle}^\text{res})$, (e) $\mathcal{L}_\text{SR} + \lambda(\mathcal{L}_\text{photo}^\text{res} + \mathcal{L}_\text{smooth} + \mathcal{L}_\text{cycle}^\text{res})$, (f)  $\mathcal{L}_\text{SR} + \lambda(\mathcal{L}_\text{photo}^\text{res} + \mathcal{L}_\text{smooth} + \mathcal{L}_\text{cycle}^\text{res} + \mathcal{L}_\text{cons}^\text{res})$.}\label{FigLoss}
 \vspace{-0.2cm}
 \end{figure}

\textbf{Losses.}\label{SecAblationLosses}
 We retrained our network using different losses to validate their effectiveness. As shown in Table~\ref{TabLoss}, the PSNR value of our network is decreased from 25.615 to 25.527 if only the SR loss is considered. That is, our network cannot well incorporate cross-view information without using the additional losses for regularization. In contrast, the SR performance is gradually improved if the photometric loss, cycle loss, smoothness loss, and stereo consistency loss are added. Note that, a 0.159 dB PSNR improvement is introduced when the network is trained with these losses calculated on residual images. As demonstrated in Section~\ref{SecLosses}, by applying these residual losses, the illuminance changes between stereo images can be eliminated and the high-frequency texture regions can be focused on.

 Moreover, we visualize the attention maps of scene \textit{Motorcycle} and \textit{Sword2} \cite{Mid} in Fig.~\ref{FigLoss}. It can be observed that the attention maps trained only with the SR loss suffer from heavy noise (Fig.~\ref{FigLoss} (a1)) and missing correspondence (Fig.~\ref{FigLoss} (a2)). When the residual photometric loss is introduced, the noise can be reduced but the problem of missing correspondence cannot be handled. That is because, the initial score map $\mathbf{S}$ has similar values at different locations in textureless regions (e.g., regions marked by the blue stroke in scene \textit{Sword2}). Consequently, a single point in the left view can be correlated to a number of points along the epipolar line in the right view, resulting in ambiguities in attention maps. When the smoothness loss is added, noise can be eliminated but the problem of missing correspondence becomes more severe (Figs.~\ref{FigLoss}(c1) and (c2)). In contrast, if the residual cycle loss is added, the missing correspondence problem can be handled but the noise cannot be reduced (Fig.~\ref{FigLoss}(d1)). This problem can be handled by introducing both smoothness loss and residual cycle loss (Figs.~\ref{FigLoss} (e1) and (e2)). Finally, the proposed residual stereo consistency loss can further enhance the stereo consistency to produce accurate and reasonable attention maps.

 \begin{table}[t]
\centering
\scriptsize
\caption{Results achieved on the \textit{KITTI 2015} dataset by iPASSR with different settings in biPAM for $4\times$SR. Here, PSNR$/$SSIM values of the cropped left images (\ie, \textit{Left}) and a pair of stereo images (\ie, $\left(\textit{Left}+\textit{Right}\right)/2$) are reported.} \label{TabbiPAM}
 \vspace{-0.2cm}
\begin{tabular}{|l|c|c|}
\hline
Models & \textit{Left} & $\left(\textit{Left}+\textit{Right}\right)/2$\\
\hline
\textit{iPASSR w/o whiten layer}                         & 25.535$/$0.7830 & 26.125$/$0.8037 \\
\textit{iPASSR w/o using valid mask}                  & 25.574$/$0.7843 & 26.179$/$0.8051 \\
\textit{iPASSR}                                                  & 25.615$/$0.7850 & 26.316$/$0.8084 \\
\hline
\end{tabular}
 \vspace{-0.1cm}
\end{table}

\textbf{Whiten layer.}
We validate the effectiveness of whiten layers by removing them from our biPAM (\ie, \textit{iPASSR w/o whiten layer}). As shown in Table~\ref{TabbiPAM}, the average PSNR value suffers a decrease of 0.191 dB if whiten layers are removed. That is because, the whiten layers can help to generate robust pairwise correspondence which is beneficial to stereo image SR.

\textbf{Valid mask.}
We demonstrate the effectiveness of valid mask by removing it from both our network and losses (\ie, \textit{iPASSR w/o valid mask}). That is, the converted features in biPAM are directly concatenated with the original features on the target side. Meanwhile, all the losses are applied equally to all spatial locations without considering occlusions. It can be observed in Table~\ref{TabbiPAM} that the average PSNR value suffers a decrease of 0.137 dB (26.179 v.s. 26.316) if the valid mask is not used.

\begin{table*}
\centering
\scriptsize
 \vspace{-0.0cm}
\caption{Quantitative results achieved by different methods for 2$\times$ and 4$\times$SR. \textit{\#Params.} represents the number of parameters of the networks. Here, PSNR/SSIM values achieved on both the cropped left images (\ie, \textit{Left}) and a pair of stereo images (\ie, $\left(\textit{Left}+\textit{Right}\right)/2$) are reported. The best results are in \textcolor{red}{red} and the second best results are in \textcolor{blue}{blue}.} \label{TabQuantitative}
 \vspace{-0.2cm}
\begin{tabular}{|l|c|c|ccc|cccc|}
\hline
\multirow{2}*{Method} & \multirow{2}*{\textit{Scale}} & \multirow{2}*{\textit{\#Params.}} & \multicolumn{3}{c|}{\textit{Left}} & \multicolumn{4}{c|}{$\left(\textit{Left}+\textit{Right}\right)/2$}\\
\cline{4-10}
         &      &           & \textit{KITTI 2012} & \textit{KITTI 2015} & \textit{Middlebury} & \textit{KITTI 2012} & \textit{KITTI 2015} & \textit{Middlebury} & \textit{Flickr1024}\\
\hline
\hline
Bicubic                                  &2$\times$ &   ---    & 28.44$/$0.8808 & 27.81$/$0.8814  & 30.46$/$0.8979 & 28.51$/$0.8842 & 28.61$/$0.8973 & 30.60$/$0.8990 & 24.94$/$0.8186 \\
 \textit{VDSR} \cite{VDSR}              &2$\times$ & 0.66M & 30.17$/$0.9062 & 28.99$/$0.9038 & 32.66$/$0.9101 & 30.30$/$0.9089 & 29.78$/$0.9150& 32.77$/$0.9102 & 25.60$/$0.8534\\
 \textit{EDSR} \cite{EDSR}              &2$\times$ & 38.6M & 30.83$/$0.9199 & 29.94$/$\textcolor{blue}{0.9231} & \textcolor{blue}{34.84}$/$\textcolor{red}{0.9489} &\textcolor{blue}{30.96}$/$\textcolor{blue}{0.9228} & 30.73$/$0.9335 & \textcolor{red}{34.95}$/$\textcolor{red}{0.9492} & \textcolor{red}{28.66}$/$\textcolor{blue}{0.9087} \\
 \textit{RDN} \cite{RDN}                   &2$\times$ & 22.0M  & 30.81$/$0.9197 & 29.91$/$0.9224 & \textcolor{red}{34.85}$/$\textcolor{blue}{0.9488} &30.94$/$0.9227 & 30.70$/$0.9330 & \textcolor{blue}{34.94}$/$\textcolor{blue}{0.9491} & \textcolor{blue}{28.64}$/$0.9084 \\
 \textit{RCAN} \cite{RCAN}            &2$\times$ & 15.3M & \textcolor{blue}{30.88}$/$\textcolor{blue}{0.9202} & \textcolor{blue}{29.97}$/$\textcolor{blue}{0.9231} & 34.80$/$0.9482 & 31.02$/$0.9232 & \textcolor{blue}{30.77}$/$\textcolor{blue}{0.9336} & 34.90$/$0.9486 & 28.63$/$0.9082 \\
 \textit{StereoSR} \cite{StereoSR}    &2$\times$ &1.08M & 29.42$/$0.9040 & 28.53$/$0.9038 & 33.15$/$0.9343 & 29.51$/$0.9073 & 29.33$/$0.9168 & 33.23$/$0.9348 & 25.96$/$0.8599 \\
 \textit{PASSRnet} \cite{PASSRnet} &2$\times$ & 1.37M & 30.68$/$0.9159 & 29.81$/$0.9191 & 34.13$/$0.9421 & 30.81$/$0.9190 & 30.60$/$0.9300 & 34.23$/$0.9422 & 28.38$/$0.9038 \\
 \textit{iPASSR} (ours)                     &2$\times$ & 1.37M &\textcolor{red}{30.97}$/$\textcolor{red}{0.9210} & \textcolor{red}{30.01}$/$\textcolor{red}{0.9234} & 34.41$/$0.9454 & \textcolor{red}{31.11}$/$\textcolor{red}{0.9240} & \textcolor{red}{30.81}$/$\textcolor{red}{0.9340} & 34.51$/$0.9454 & 28.60$/$\textcolor{red}{0.9097} \\
\hline
\hline
Bicubic                                    &4$\times$  &   ---     & 24.52$/$0.7310 & 23.79$/$0.7072 & 26.27$/$0.7553 & 24.58$/$0.7372 & 24.38$/$0.7340 & 26.40$/$0.7572 & 21.82$/$0.6293 \\
 \textit{VDSR} \cite{VDSR}               & 4$\times$ & 0.66M & 25.54$/$0.7662 & 24.68$/$0.7456 & 27.60$/$0.7933 & 25.60$/$0.7722 & 25.32$/$0.7703 & 27.69$/$0.7941 & 22.46$/$0.6718 \\
 \textit{EDSR} \cite{EDSR}              & 4$\times$ & 38.9M & 26.26$/$0.7954 & 25.38$/$0.7811 & \textcolor{blue}{29.15}$/$\textcolor{blue}{0.8383} & 26.35$/$0.8015 & 26.04$/$0.8039 & 29.23$/$\textcolor{blue}{0.8397} & 23.46$/$0.7285 \\
 \textit{RDN} \cite{RDN}                  & 4$\times$ & 22.0M  & 26.23$/$0.7952 & 25.37$/$0.7813 & \textcolor{blue}{29.15}$/$\textcolor{red}{0.8387} & 26.32$/$0.8014 & 26.04$/$0.8043 & \textcolor{blue}{29.27}$/$\textcolor{red}{0.8404} & \textcolor{blue}{23.47}$/$\textcolor{red}{0.7295} \\
 \textit{RCAN} \cite{RCAN}               & 4$\times$ & 15.4M & \textcolor{blue}{26.36}$/$\textcolor{blue}{0.7968} & 25.53$/$\textcolor{blue}{0.7836} & \textcolor{red}{29.20}$/$0.8381 & \textcolor{blue}{26.44}$/$\textcolor{blue}{0.8029} & \textcolor{blue}{26.22}$/$\textcolor{blue}{0.8068} & \textcolor{red}{29.30}$/$\textcolor{blue}{0.8397} & \textcolor{red}{23.48}$/$0.7286 \\
 \textit{PASSRnet}  & 4$\times$ & 1.42M   & 26.26$/$0.7919 & 25.41$/$0.7772 &28.61$/$0.8232 & 26.34$/$0.7981 & 26.08$/$0.8002 & 28.72$/$0.8236 & 23.31$/$0.7195 \\
 \textit{SRRes+SAM} \cite{SAM}    & 4$\times$ & 1.73M  & 26.35$/$0.7957 & \textcolor{blue}{25.55}$/$0.7825 & 28.76$/$0.8287 & \textcolor{blue}{26.44}$/$0.8018 & \textcolor{blue}{26.22}$/$0.8054 & 28.83$/$0.8290 & 23.27$/$0.7233 \\
 \textit{iPASSR} (ours)                      & 4$\times$ & 1.42M  & \textcolor{red}{26.47}$/$\textcolor{red}{0.7993} & \textcolor{red}{25.61}$/$\textcolor{red}{0.7850} & 29.07$/$0.8363 & \textcolor{red}{26.56}$/$\textcolor{red}{0.8053} & \textcolor{red}{26.32}$/$\textcolor{red}{0.8084} & 29.16$/$0.8367 & 23.44$/$\textcolor{blue}{0.7287} \\
\hline
\end{tabular}
\begin{tabular}{l}
\leftline{Note: We do not present $2\times$SR results of \textit{SRRes+SAM} \cite{SAM} and $4\times$SR results of \textit{StereoSR} \cite{StereoSR} since their models are unavailable.}\\
\end{tabular}
 \vspace{-0.3cm}
\end{table*}

\begin{figure}
  \centering
  \includegraphics[width=8.3cm]{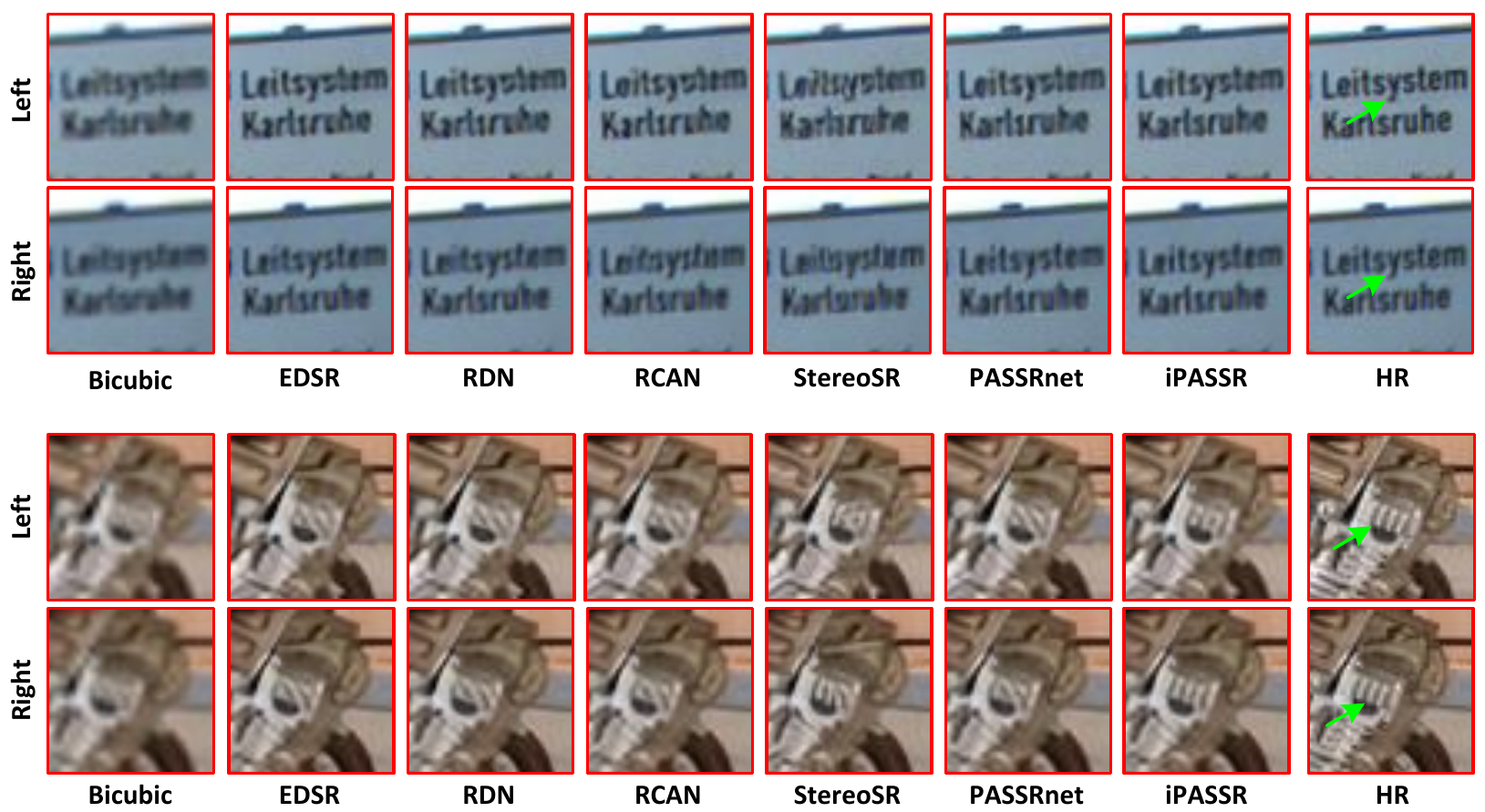}
  \vspace{-0.4cm}
  \caption{Visual results ($2\times$) achieved by different methods on the \textit{KITTI 2015} (top) and \textit{Middlebury} datasets (bottom).}\label{Fig2xSR}
  \vspace{-0.1cm}
\end{figure}

\subsection{Comparison to state-of-the-arts methods}
In this section, we compare our \textit{iPASSR} to several state-of-the-art methods, including four SISR methods \ie, \textit{VDSR} \cite{VDSR}, \textit{EDSR} \cite{EDSR}, \textit{RDN} \cite{RDN}, \textit{RCAN} \cite{RCAN}) and three stereo image SR methods\footnote{We do not compare our method to \textit{SPAMnet} \cite{SPAM} and \textit{DASSR} \cite{DASSR} because: \textbf{(1)} their codes and models are unavailable, \textbf{(2)} The evaluation schemes in \cite{SPAM,DASSR} are different from those in \cite{StereoSR,PASSRnet,SAM}, so that we cannot directly copy the PSNR and SSIM scores in their papers.} (\ie, \textit{StereoSR} \cite{StereoSR}, \textit{PASSRnet} \cite{PASSRnet}, \textit{SRRes+SAM} \cite{SAM}). Note that, we retrained all SISR methods \cite{VDSR,EDSR,RDN,RCAN} on our training set for fair comparison.

\begin{figure}
  \centering
  \includegraphics[width=8.3cm]{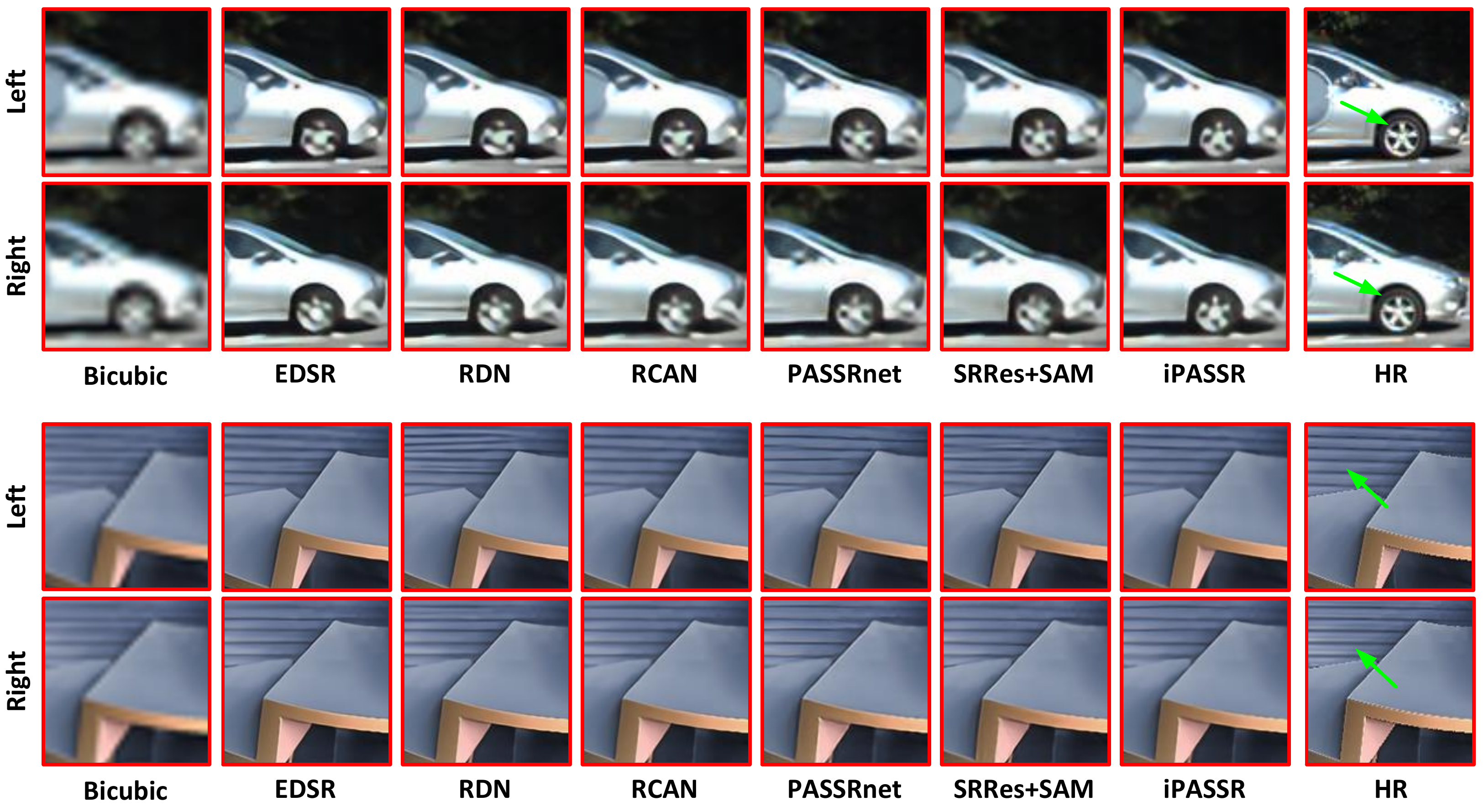}
  \vspace{-0.5cm}
  \caption{Visual results ($4\times$) achieved by different methods on the \textit{KITTI 2015} (top) and \textit{Flickr1024} datasets (bottom).}\label{Fig4xSR}
  \vspace{-0.1cm}
\end{figure}

 \textbf{Quantitative results.}
   As shown in Table~\ref{TabQuantitative}, our \textit{iPASSR} achieves the highest PSNR and SSIM values on the \textit{KITTI 2012} and \textit{KITTI 2015} datasets for $2\times$ and $4\times$ SR. For the \textit{Middlebury} and \textit{Flickr1024} datasets, our \textit{iPASSR} outperforms all stereo image SR methods, but is slightly inferior to \textit{EDSR}, \textit{RDN}, and \textit{RCAN}. Note that, the model sizes of our \textit{iPASSR} are comparable to \textit{PASSRnet} but significantly smaller than \textit{EDSR}, \textit{RDN} and \textit{RCAN}\footnote{It is worth noting that \textit{DRCN} \cite{DRCN}, \textit{DRRN} \cite{DRRN} and \textit{LapSRN} \cite{lapSRN} which have comparable number of parameters as our \textit{iPASSR} were not included for comparison since they have already been outperformed by \textit{PASSRnet} as demonstrated in \cite{PASSRnet}. In this paper, we investigate the performance gap between our method and the top-performing SISR methods \cite{EDSR,RDN,RCAN}, which is the first attempt in this area. We hope these comparative results can inspire the future research of stereo image SR.}. Although a large model enables rich and hierarchical feature representation which can boost the SR performance, we decided to keep our \textit{iPASSR} lightweight and improve SR performance by exploiting cross-view information in stereo images.

  %It is also worth noting that, all methods in Table~\ref{TabQuantitative} have not been trained on the KITTI 2012 \cite{K12} and KITTI 2015 \cite{K15} datasets. However, our iPASSR achieves superior performance on these two datasets. It not only demonstrates the generalization capability of our iPASSR, but also demonstrates the importance of cross-view information when SR methods are applied to a new dataset.

%It is worth noting that, although it is reported in Table~\ref{TabQuantitative} that our iPASSR have comparable FLOPs as compared to PASSRnet, our iPASSR is significantly more efficient and has lower computational cost. That is because, PASSRnet super-resolves only the left image while our iPASSR achieves SR on both side of views. Consequently, features of both sides need to pass through our reconstruction module, resulting in 18.9 G and 19.9 G additional FLOPs for $2\times$ and $4\times$SR, respectively. Since PASSRnet performs parallax correlation twice in its PAM to generate attention maps of both sides while our proposed biPAM only needs to perform parallax correlation once (i.e., Eq.~\ref{EqScore}) to generate attention maps of both sides, which is more computational efficient.

 \textbf{Qualitative results.}
 Qualitative results for $2\times$ and $4\times$ SR are shown in Figs.~\ref{Fig2xSR} and \ref{Fig4xSR}, respectively. Readers can view this \href{https://wyqdatabase.s3-us-west-1.amazonaws.com/iPASSR_visual_comparison.mp4}{demo video} for better comparison. Since input LR images are degraded by the downsampling operation, the SR process is highly ill-posed especially for $4\times$SR. In such cases, SISR methods only use spatial information and cannot well recover the missing details. In contrast, our \textit{iPASSR} use cross-view information to produce more faithful details with fewer artifacts. Moreover, the images generated by our \textit{iPASSR} are more stereo-consistent than those generated by \textit{PASSRnet} and \textit{SRRes+SAM}.

\begin{figure}
 \centering
 \includegraphics[width=8.3cm]{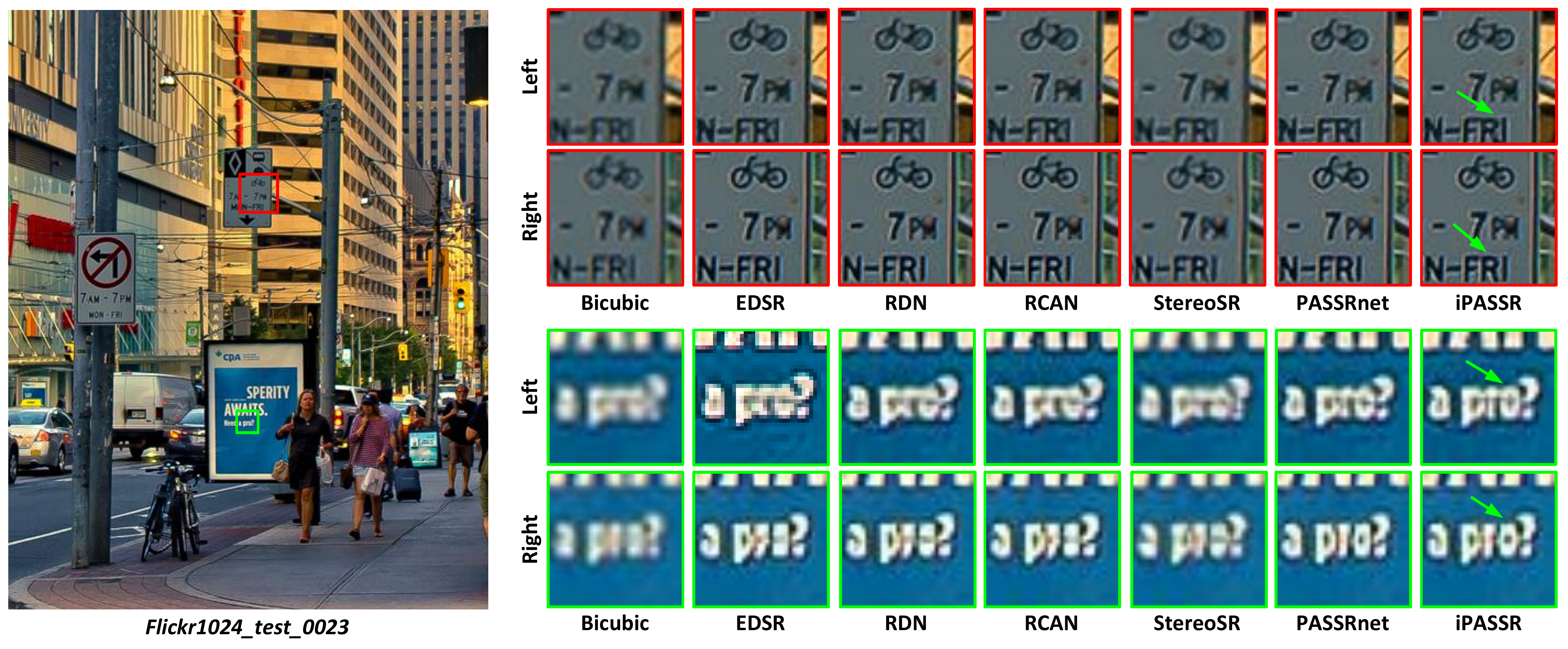}
  \vspace{-0.5cm}
 \caption{Visual results achieved by different methods on real-world images \cite{Flickr} for $2\times$SR.}\label{FigRealSR}
 \vspace{-0.3cm}
 \end{figure}

\textbf{Performance on real-world images.}
We test the performance of different methods on real-world stereo images by directly applying them to an HR image pair from the Flickr1024 dataset \cite{Flickr}. As shown in Fig.~\ref{FigRealSR}, our \textit{iPASSR} achieves better perceptual quality than the compared methods.  It is worth noting that, left and right views of an image pair may suffer different degrees of degradation in real-world cases (e.g., in the region marked by the red box, the left image suffers more severe blurs than the right one). SISR methods cannot well recover the missing details by using the intra-view information only. In contrast, our \textit{iPASSR} benefits from the cross-view information and produce images with less blurring artifacts.
 %Consequently, state-of-the-art SISR methods \cite{EDSR,RDN,RCAN} can not well recover the characters in the left view based on its intra-view information only. In contrast, our method can handle this challenging issue by using the complementary cross-view information from the other view.
 %Since the horizontal epipolar constraint between stereo image pairs holds under both bicubic and real-world degradation, our iPASSR can learn to incorporate cross-view information from training stereo images. These experimental results also demonstrated that our iPASSR can be easily applied to stereo cameras to generate high-quality HR images.

 \begin{table}
\centering
 \scriptsize
\caption{Quantitative results achieved by \textit{GwcNet} \cite{GwcNet} on 4$\times$ SR stereo images. All these metrics were averaged on the test set of the \textit{SceneFlow} dataset \cite{SceneFlow}, where lower values indicate better performance. Best results are in \textcolor{red}{red} and the second best results are \textcolor{blue}{blue}.} \label{TabDisp}
\vspace{-0.2cm}
\begin{tabular}{|l|c|c|c|c|}
\hline
Method & ~~\textit{EPE}~~  & \textit{$>$1px (\%)} & \textit{$>$2px (\%)} & \textit{$>$3px (\%)} \\
\hline
Bicubic                                 & 1.196     & 11.5 & 5.96 & 4.28 \\
VDSR \cite{VDSR}              & 1.068     & 10.8 & 5.37 & 3.80 \\
PASSRnet \cite{PASSRnet,PAM}  & 1.019     & 11.5 & 5.44 & 3.72 \\
SRRes+SAM \cite{SAM}     & 0.991     & 11.1 & 5.18 & 3.57 \\
iPASSR (ours)                      & \textcolor{blue}{0.949}     & \textcolor{blue}{10.0} & \textcolor{blue}{4.79} & \textcolor{blue}{3.35} \\
HR                                       & \textcolor{red}{0.667}    & \textcolor{red}{6.77} & \textcolor{red}{3.34} & \textcolor{red}{2.38} \\
\hline
\end{tabular}
\vspace{0.1cm}
\end{table}

 \begin{figure}
 \centering
 \includegraphics[width=8.3cm]{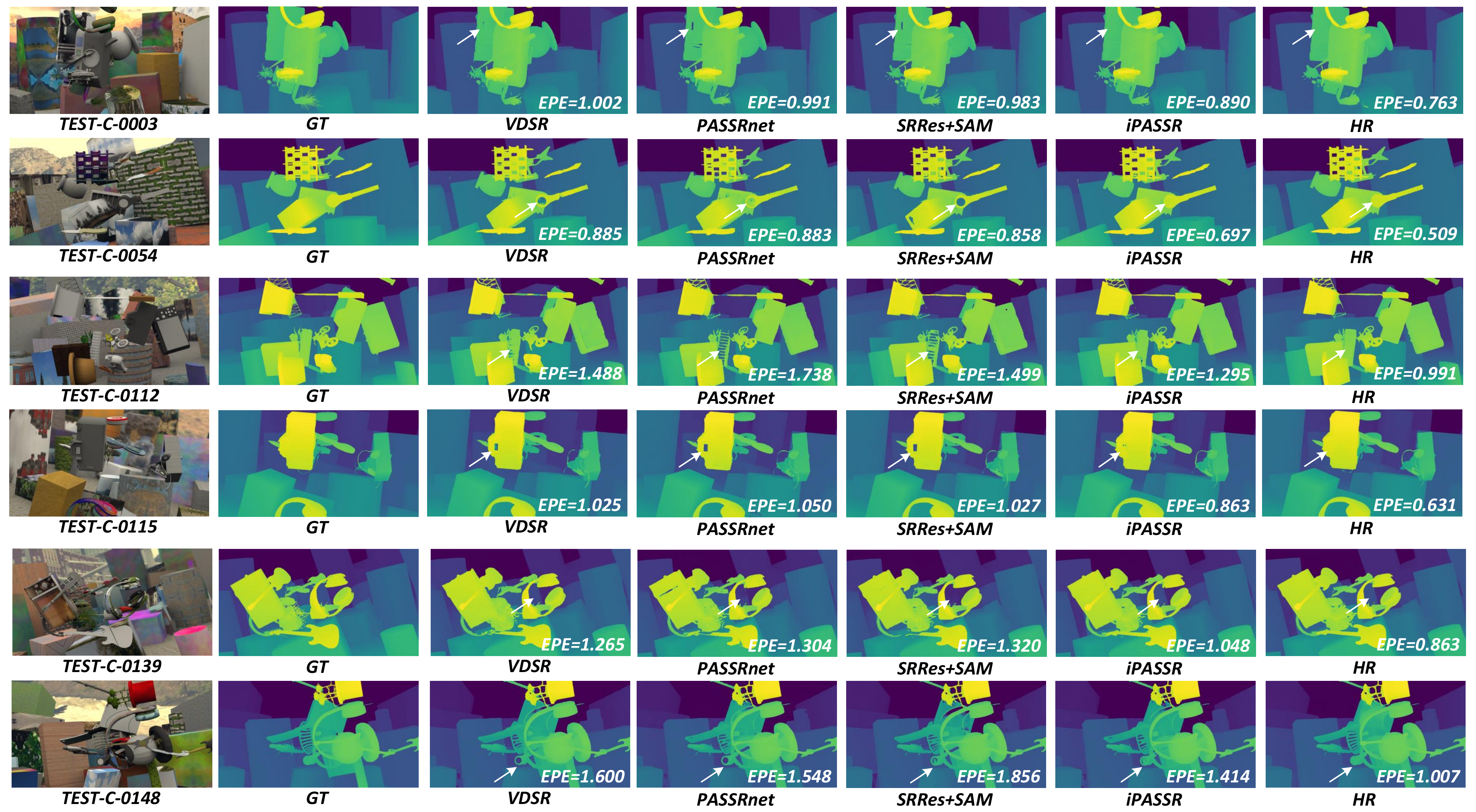}
  \vspace{-0.3cm}
 \caption{Qualitative results achieved by \textit{GwcNet} \cite{GwcNet} using $4\times$SR stereo images generated by different SR methods.}\label{FigDisp}
 \vspace{-0.2cm}
 \end{figure}

 \textbf{Benefits to disparity estimation.}
 As stereo-consistent and HR image pairs are beneficial to disparity estimation, we investigate this benefit by using the super-resolved stereo images for disparity estimation. We performed 4$\times$ downsampling on the images from the test sets of the \textit{SceneFlow} dataset\footnote{All 145 scenes under path ``./TEST/C/'' were used as the test set in this paper. For stereo images of each scene, only the first frame (\ie, ``./left/0006.png'' and ``./right/0006.png'') was used.} \cite{SceneFlow}, and used different methods to super-resolve these LR images to their original resolution. Then, we applied \textit{GwcNet} \cite{GwcNet} to these super-resolved stereo images for disparity estimation. The original HR images and bicubicly upsampled images were used to produce the upper bound and the baseline results, respectively. End-point-error (EPE) and t-pixel error rate ($>\textit{tpx}$) were used as quantitative metrics to evaluate the estimated disparity. As shown in Table~\ref{TabDisp}, a 0.529 (\ie, $79.3\%$) increase in EPE is introduced when HR input images are replaced with the bicubicly interpolated ones. It demonstrates that the details (e.g., edges and textures) in the stereo images are important to disparity estimation. Note that, our \textit{iPASSR} can better reduce the error by providing high-quality and stereo-consistent stereo images. The visual examples in Fig~\ref{FigDisp} demonstrate that the disparity map corresponding to our method is more accurate and close to the one estimated from HR stereo images.

\subsection{Discussion}\label{Discussion}
 During the retraining of SISR methods, we noticed that the training dataset has an influence on the SR performance. To investigate the influence of training datasets, we used \textit{EDSR} and \textit{RCAN} developed on different datasets to perform stereo image SR. As shown in Table~\ref{TabTrainset}, \textit{EDSR} and \textit{RCAN} achieve better performance when trained on the \textit{DIV2K} dataset \cite{NTIRE2017}. That is because, the \textit{DIV2K} dataset was specifically developed for SISR and has higher-quality images than existing stereo image datasets. To demonstrate this claim, we use three no-reference image quality assessment metrics \cite{BRISQUE,NIQE,CEIQ} to evaluate the image quality of these datasets. As shown in Table~\ref{DatasetComparison}, the \textit{DIV2K} dataset achieves the best results in terms of all the metrics. It demonstrates that high-quality training images can introduce a notable performance gain to deep SR networks.

\begin{table}
\centering
\scriptsize
\caption{Comparative results achieved by \textit{EDSR} and \textit{RCAN} with different training sets for both $2\times$ and $4\times$SR.} \label{TabTrainset}
\vspace{-0.2cm}
\begin{tabular}{|p{1.3cm}<{\centering}| p{0.2cm}<{\centering}| p{1.05cm}<{\centering} p{1.05cm}<{\centering} p{1.05cm}<{\centering} p{1.05cm}<{\centering}|}
\hline
Method  &    & \textit{KITTI2012} & \textit{KITTI2015} & \textit{Middlebury} & \textit{Flickr1024} \\
\hline
\rowcolor{shadow}
\textit{EDSR\_div2k}     & 2$\times$  & 31.06/0.925 & 30.77/0.935 & 35.34/0.951 & 28.58/0.909 \\
\textit{EDSR\_stereo}    & 2$\times$ & 30.95/0.923 & 30.73/0.934 & 34.95/0.949& 28.66/0.908 \\
\rowcolor{shadow}
\textit{RCAN\_div2k}    & 2$\times$ & 31.16/0.926 & 30.88/0.945 & 35.42/0.952 & 28.64/0.910 \\
\textit{RCAN\_stereo}   & 2$\times$& 31.02/0.923 & 30.77/0.934 & 34.90/0.949 & 28.63/0.908 \\
\hline
\rowcolor{shadow}
\textit{EDSR\_div2k}  &4$\times$  & 26.62/0.809 & 26.39/0.814 & 29.48/0.842 & 23.58/0.735 \\
\textit{EDSR\_stereo}  &4$\times$   & 26.35/0.802 & 26.04/0.804 & 29.23/0.840 & 23.46/0.729\\
\rowcolor{shadow}
\textit{RCAN\_div2k}  &4$\times$  & 26.65/0.809 & 26.45/0.814 & 29.56/0.845 & 23.60/0.737 \\
\textit{RCAN\_stereo} & 4$\times$   & 26.44/0.803 & 26.22/0.807 & 29.30/0.840 & 23.48/0.729 \\
\hline
\end{tabular}
\vspace{-0.0cm}
\end{table}

\begin{table}
\centering
 \scriptsize
\caption{No-reference perceptual quality scores of different SR datasets. Both the average value and the standard deviation are reported. Lower scores of \textit{BRISQUE} \cite{BRISQUE}, \textit{NIQE} \cite{NIQE} and higher scores of \textit{CEIQ} \cite{CEIQ} indicate better quality.} \label{DatasetComparison}
\vspace{-0.2cm}
\begin{tabular}{|l|ccc|}
\hline
Dataset & BRISQUE ($\downarrow$) & NIQE  ($\downarrow$) & CEIQ  ($\uparrow$) \\
\hline
\textit{KITTI 2012}        &  17.30 ($\pm$ 6.60)  &  3.22 ($\pm$0.42)       & 3.31 ($\pm$0.14)   \\
\textit{KITTI 2015}        &  26.41 ($\pm$ 5.26)  &  3.23 ($\pm$0.48)       &  3.34 ($\pm$0.19)  \\
\textit{Middlebury}        &  14.88 ($\pm$ 9.19)  &  3.77 ($\pm$0.99)       & 3.31 ($\pm$0.21)  \\
\textit{Flickr1024}         &  19.10 ($\pm$13.57)  &  3.40 ($\pm$0.99)     & 3.25 ($\pm$0.36)  \\
\textit{DIV2K}               &  11.40 ($\pm$11.98)  &  2.99 ($\pm$1.05)     & 3.36 ($\pm$0.30)  \\
\hline
\end{tabular}
\vspace{-0.1cm}
\end{table}

\section{Conclusion}\label{SecConclusion}
In this paper, we proposed a method to exploit symmetry cues for stereo image SR. We first proposed a bi-directional parallax attention module (biPAM) and an inline occlusion handling scheme to effectively interact cross-view information, and then equipped biPAM to a Siamese network to develop our \textit{iPASSR}. Moreover, we proposed several residual losses to achieve robustness to illuminance changes. Extensive ablation studies were performed to validate the effectiveness of our design choices, and comparative results on four public datasets demonstrated the state-of-the-art performance of our method. Furthermore, we made an in-depth analysis on the benefits of stereo image SR to disparity estimation, and the influence of training datasets to image SR.

\section*{Acknowledgement}
This work was partially supported in part by the National Natural Science Foundation of China (Nos. 61972435, 61401474, 61921001).
%As discussed in Section \ref{Discussion}, high-quality training images can introduce significant performance improvements to SR methods by enhancing their intra-view information exploitation capability. However, these high-quality single image datasets such as DIV2K \cite{NTIRE2017} cannot be directly used to train stereo image SR methods. Therefore, it is still an open issue to make stereo image SR methods more powerful in intra-view information exploitation. To handle this problem, a straight-forward strategy is to develop a high-quality dataset specifically designed for stereo image SR. However, developing large-scale datasets with diverse scenarios under stereo settings is labor-intensive. Another solution is to design novel algorithms that can effectively learn natural context priors from an SISR dataset for stereo image SR. In the future, we will try to further improve the performance of stereo image SR methods.

 %For instance,  has been recently demonstrated effective for efficient image SR. When performing knowledge distillation for stereo image SR, powerful SISR networks trained on the high-quality single image datasets can be set as the teacher network to ``teach'' the student stereo image SR networks. In this way, the informative context priors can be gradually distilled to improve the intra-view information exploitation capability of student networks. In the future, we will investigate the aforementioned potential solutions to further improve the performance of stereo image SR methods.
{\small
\bibliographystyle{ieee_fullname}
\bibliography{iPASSR}
}

\end{document}